\documentclass[preprint,3p,twocolumn]{elsarticle}

\usepackage{amssymb}
\usepackage{amsmath}

\usepackage{pifont}
\usepackage{xcolor}
\usepackage{multirow}
\usepackage{tabularx}
\usepackage{booktabs}
\usepackage{colortbl}
\usepackage{subcaption}

\journal{Neurocomputing}

\begin{document}

\begin{frontmatter}



\title{DynaMix: Generalizable Person Re-identification\\via Dynamic Relabeling and Mixed Data Sampling}

\author[label1,label2]{Timur Mamedov}
\author[label2]{Anton Konushin}
\author[label1]{Vadim Konushin}

\affiliation[label1]{organization={Tevian},
            city={Moscow},
            country={Russia}}

\affiliation[label2]{organization={Lomonosov Moscow State University},
            country={Russia}}

\begin{abstract}
Generalizable person re-identification (Re-ID) aims to recognize individuals across unseen cameras and environments. While existing methods rely heavily on limited labeled multi-camera data, we propose DynaMix, a novel method that effectively combines manually labeled multi-camera and large-scale pseudo-labeled single-camera data. Unlike prior works, DynaMix dynamically adapts to the structure and noise of the training data through three core components: (1) a Relabeling Module that refines pseudo-labels of single-camera identities on-the-fly; (2) an Efficient Centroids Module that maintains robust identity representations under a large identity space; and (3) a Data Sampling Module that carefully composes mixed data mini-batches to balance learning complexity and intra-batch diversity. All components are specifically designed to operate efficiently at scale, enabling effective training on millions of images and hundreds of thousands of identities. Extensive experiments demonstrate that DynaMix consistently outperforms state-of-the-art methods in generalizable person Re-ID.
\end{abstract}




\end{frontmatter}



\section{Introduction}
\label{sec:intro}

Person re-identification (Re-ID) aims to recognize individuals across multiple non-overlapping cameras. This task is essential for various real-world applications, such as surveillance and security systems. While modern methods achieve high accuracy within specific datasets (standard person Re-ID), they often struggle to generalize to novel cameras and unseen environments (generalizable person Re-ID).

To learn discriminative features from various viewpoints, Re-ID methods need to be trained using data from multiple cameras (multi-camera data). However, collecting and labeling such data requires much effort, leading to relatively small and limited multi-camera datasets (Tab.~\ref{tab:datasets}). This constraint reduces the diversity of training environments and hinders the generalization ability of Re-ID models when deployed in the wild.

\begin{figure}[t]
    \centering
    \includegraphics[width=0.792\linewidth]{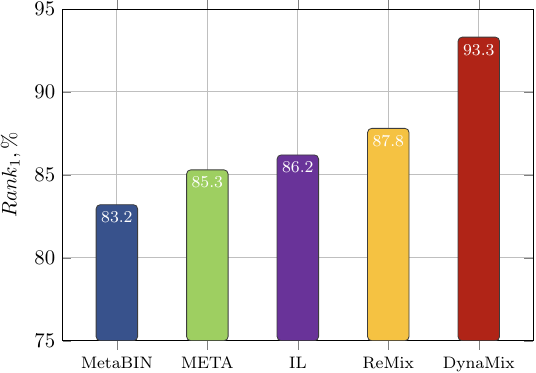}
    \caption{$Rank_1$ scores of DynaMix and other state-of-the-art methods in the multi-source cross-dataset scenario~(D+C3+MS $\rightarrow$ M  configuration in Tab.~\ref{tab:comparison-multi}).}
    \label{fig:comparison_graph}
\end{figure}

In contrast, images of people from one camera~(single-camera data) can be easily collected automatically, e.g., from YouTube videos \cite{fu2021unsupervised, fu2022large}, capturing numerous diverse identities across various locations (Tab.~\ref{tab:datasets}). Yet, since each individual is captured by only one camera and viewpoint, the model is not essentially encouraged to decouple identity features and camera features, which may impose limitations on the generalization ability. Therefore, directly adding such simple single-camera data to the training process worsens Re-ID performance~\cite{mamedov2025remix} and most existing approaches use it only for self-supervised pre-training \cite{fu2021unsupervised, fu2022large, mamedov2023approaches}.

Recent ReMix \cite{mamedov2025remix} shows that single-camera data can be directly used together with multi-camera data during training, leading to improved generalization. However, it leaves two key issues unresolved: (1) noisy pseudo-labels in large single-camera datasets are used without refinement, causing incorrect supervision; and (2) data sampling is entirely random, ignoring the strong distributional differences between multi-camera and single-camera data. These limitations prevent ReMix from fully exploiting the scale and diversity of single-camera data.

We propose DynaMix, a novel generalizable person Re-ID method built upon the high-level idea of mixed data training but introducing three new modules designed specifically to overcome the limitations of ReMix. Unlike previous works, DynaMix dynamically adapts to noisy labels, scales to hundreds of thousands of identities, and intelligently mixes heterogeneous data during training.

Our DynaMix consists of three tightly coupled components:
\begin{itemize}\vspace{-6pt}
    \setlength\itemsep{-0.1em}
    \item a Relabeling Module that dynamically refines noisy pseudo-labels in single-camera data through on-the-fly filtering, reassignment, and identity merging;
    \item an Efficient Centroids Module that enables scalable identity representation updates under hundreds of thousands of pseudo-labeled identities;
    \item a Data Sampling Module that carefully composes mixed data mini-batches to balance learning complexity and intra-batch diversity.
\end{itemize}\vspace{-6pt}
These components work in synergy to produce a strong and scalable Re-ID model capable of generalizing to novel cameras and unseen environments. Extensive experiments across all cross-dataset and multi-source cross-dataset benchmarks demonstrate that DynaMix significantly outperforms existing state-of-the-art methods in generalizable person Re-ID~(Fig.~\ref{fig:comparison_graph}).

\begin{figure*}[t]
    \centering
    \includegraphics[width=0.753\linewidth]{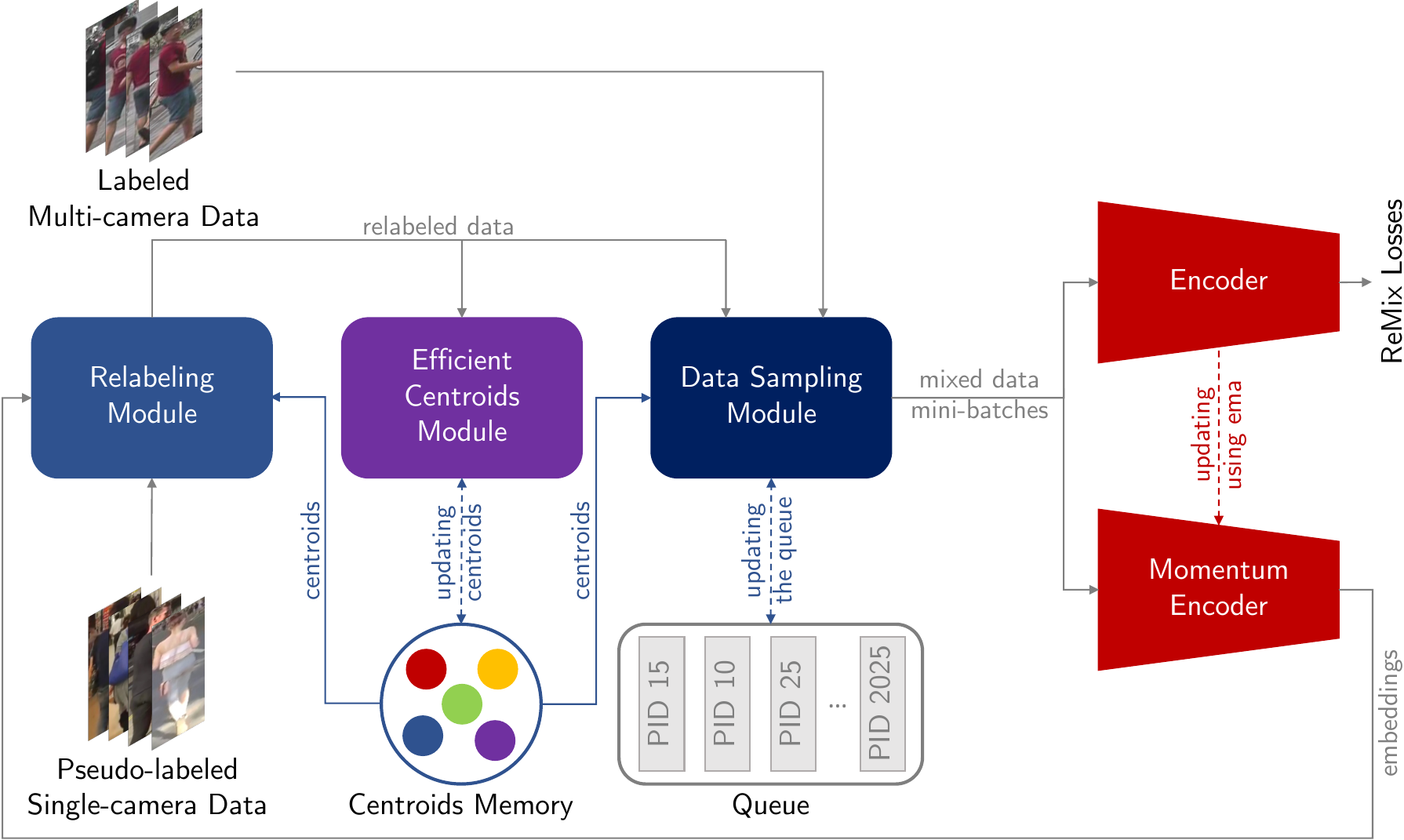}
    \caption{Overview of DynaMix. Our method leverages both labeled multi-camera and pseudo-labeled single-camera data during training. It consists of three mutually dependent components: the Relabeling Module dynamically refines noisy pseudo-labels; the Efficient Centroids Module enables scalable identity representation updates; and the queue-based Data Sampling Module composes mixed data mini-batches to balance learning complexity and intra-batch diversity.}
    \label{fig:method-scheme}
\end{figure*}

\section{Related Work}
\label{sec:related}

\subsection{Standard Person Re-ID}

The rapid evolution of person Re-ID methods over the past few years follows the general trends in computer vision. Early approaches relied on hand-crafted features to distinguish individuals across cameras \cite{prosser2010person, zheng2012reidentification}. With the rise of deep learning, convolutional neural networks (CNNs) dominated the field \cite{wang2018learning, luo2019bag, ni2021flipreid}. Currently, the best results in Re-ID belong to vision transformers (ViTs) \cite{he2021transreid, tan2022dynamic}. Numerous works explored training models to capture discriminative and fine-grained features \cite{zhu2022dual, zhu2022pass, zhang2023pha}, while others \cite{li2021diverse, lai2021transformer, jia2022learning} combined CNNs with transformer layers to enjoy the advantages of both architectures. At the same time, many methods enhanced Re-ID accuracy by jointly modeling global and local body-part features \cite{yan2022person, tang2022person}. Recently, CLIP-ReID \cite{li2023clip} demonstrated that Re-ID performance can be boosted using CLIP-like architectures with dual modalities.

Existing methods demonstrate strong performance in the standard Re-ID scenario, with training and testing on different parts of the same dataset. However, they often fail to generalize to novel cameras and unseen environments (generalizable person Re-ID), either by design or due to the usage of small multi-camera datasets. In this work, we show that combining multi-camera and single-camera data during training can significantly improve generalization.

\subsection{Generalizable Person Re-ID}

Generalizable person Re-ID implies that a Re-ID model performs robustly across a variety of datasets and environments. One line of research focuses on making feature representations less sensitive to domain-specific variations. Specifically, works~\cite{jin2020style, choi2021meta, jiao2022dynamically} improved cross-dataset performance through normalization techniques, while PAT \cite{ni2023part} combined local features less affected by domain shifts with global visual information. Other methods attempted to resolve the generalization issue with elaborate neural architectures. OSNet \cite{zhou2019omni} proposed a residual block with multiple convolutional streams to detect features at different scales, and further improved the robustness with specialized normalization layers \cite{zhou2021learning}. TransMatcher addressed robust Re-ID with the transformer-based model \cite{liao2021transmatcher}.

While focusing on model architecture, the aforementioned methods largely overlooked the importance of data diversity. Our empirical study reveals it to be a missed opportunity: in DynaMix, increasing data diversity through large-scale single-camera datasets yields substantial performance gains and facilitates state-of-the-art results in all cross-dataset benchmarks.

\subsection{Use of Single-Camera Data}

Single-camera data can be easily obtained from sources like YouTube videos \cite{fu2021unsupervised, fu2022large}. Such data features individuals captured from a single viewpoint using a single camera, hence being less complex from the Re-ID perspective compared to multi-camera data. Consequently, it was used mainly for self-supervised pre-training \cite{mamedov2023approaches, hu2024personvit}. In this set-up, single-camera training and multi-camera training were separated. ReMix \cite{mamedov2025remix} explored joint training on a mixture of multi-camera and single-camera data, demonstrating improved generalization. However, it faced challenges such as noisy pseudo-labels and naive random sampling.

In contrast, DynaMix introduces a fundamentally different approach, combining dynamic pseudo-label refinement, efficient identity representation updates, and adaptive mixed data sampling to enable scalable and robust training across heterogeneous data.

\subsection{Other Specialized Person Re-ID Settings}

Besides generalizable person Re-ID, a number of works explored more specialized or modality-specific settings. Clothes-changing person Re-ID focuses on learning identity cues that remain stable despite significant appearance variations caused by clothing changes \cite{ding2024clothes, ding2025decoupling, ding2025person}. Another line of research investigated visible–infrared person Re-ID~\cite{pang2023cross, ren2024implicit, tang2024visible}, where the goal is to match individuals across heterogeneous sensing modalities, commonly used in low-light or nighttime surveillance scenarios. Cross-resolution person Re-ID aims to re-identify individuals captured at different image resolutions, often arising from varying distances, sensor quality, or compression settings \cite{zhang2021deep, pang2024dual}. Re-ID was also employed as a component in multi-object tracking, where identity embeddings help maintain consistent tracks across occlusions, re-entries, or long-term interruptions \cite{wojke2017simple, mamedov2021queue, zhang2022bytetrack, mamedov2022video}.

While these works addressed related challenges, they operated under fundamentally different problem settings and data modalities compared to generalizable person Re-ID. Our work instead focuses on improving generalization across unseen cameras and domains through joint training on a mixture of multi-camera and single-camera data.

\section{Proposed Method}
\label{sec:method}

The proposed DynaMix method features an encoder and a momentum encoder with identical architecture (Fig.~\ref{fig:method-scheme}) to improve robustness to pseudo-label noise and training instability. During inference, only the momentum encoder is used, as it produces more stable and reliable feature representations.

Our key innovation is three modules that are strongly interconnected and supposed to work in synergy. The training process of DynaMix involves both labeled multi-camera data and pseudo-labeled single-camera data. Given the potential noise in pseudo-labels, the Relabeling Module (Sec.~\ref{sec:relabeling-module}) dynamically refines them on-the-fly. Meanwhile, the Data Sampling Module (Sec.~\ref{sec:data-sampling-module}) carefully composes mini-batches containing both data types, balancing learning complexity and intra-batch diversity via a queue-based system. Both modules rely on centroids for each person identity (PID), which are efficiently maintained by the Efficient Centroids Module (Sec.~\ref{sec:efficient-centroids-module}) to enable scalable training with hundreds of thousands of IDs.

DynaMix combines the Instance ($\mathcal{L}_{ins}$), Augmentation ($\mathcal{L}_{aug}$), Centroids ($\mathcal{L}_{cen}$), and Camera Centroids ($\mathcal{L}_{cc}$) losses for effective training on mixed data, following the same loss weights as defined in ReMix:
\begin{equation}
    \mathcal{L} = \mathcal{L}_{ins} + \mathcal{L}_{aug} + \mathcal{L}_{cen} + 0.5 \cdot \mathcal{L}_{cc}.
    \label{eq:full-loss}
\end{equation}
The Instance Loss pulls instances toward all positives and pushes them away from all negatives in a mini-batch, promoting generalizable features. The Augmentation Loss aligns augmented views with their originals while separating them from other identities, mitigating augmentation-induced similarity shifts. The Centroids Loss encourages alignment with class centroids, and the Camera Centroids Loss further groups same-identity instances across different cameras.

During training, the encoder weights are updated directly through backpropagation, while the momentum encoder is updated using exponential moving averaging:
\begin{equation}
    \theta^{t}_{m} = {\lambda \theta^{t-1}_{m} + (1 - \lambda) \theta^{t}_{e}},
    \label{eq:method-momentum}
\end{equation}
where ${\theta^{t}_{e}}$ and ${\theta^{t}_{m}}$ are the weights of the encoder and the momentum encoder at iteration ${t}$, respectively; and ${\lambda}$ is the momentum coefficient.

\subsection{Formal Definitions}
\label{sec:formal-definitions}

We formally describe the training datasets as follows. Labeled multi-camera data is represented as $\mathcal{D}_m = \left\{ (x_i, y_i, c_i) \right\}_{i=1}^{N_m}$, where $x_i$ is the image, $y_i \in \mathcal{Y}_m = \left\{ 1, 2, \ldots, M_m \right\}$ is the person identity label~(PID), and $c_i \in \mathcal{C}_m = \left\{ 1, 2, \ldots, K_m \right\}$ is the camera ID. Similarly, pseudo-labeled single-camera data is given by $\mathcal{D}_s = \left\{ (x_i, y_i, v_i) \right\}_{i=1}^{N_s}$, where $y_i \in \mathcal{Y}_s$ is the person identity pseudo-label and $v_i \in \mathcal{V}_s$ is the video ID. In $\mathcal{D}_s$, each person appears only in a single video, ensuring no cross-video appearances within $\mathcal{D}_s$.

To effectively utilize $\mathcal{D}_s$, we maintain a Centroids Memory, denoted as $\mathcal{M}_s = \left\{ \mu_k \right\}_{k=1}^{M_s}$, where $\mu_k \in \mathbb{R}^d$ is the centroid for the $k$-th identity pseudo-label. The Centroids Memory $\mathcal{M}_s$ is dynamically updated during training using the Efficient Centroids Module (Sec.~\ref{sec:efficient-centroids-module}). We also define a similarity function between two feature vectors $f_1$ and $f_2$ as $sim(f_1, f_2) = \frac{f_1 \cdot f_2}{\|f_1\| \|f_2\|}$.

\subsection{The Relabeling Module}
\label{sec:relabeling-module}

\begin{figure}[t]
    \centering
    \includegraphics[width=0.9\linewidth]{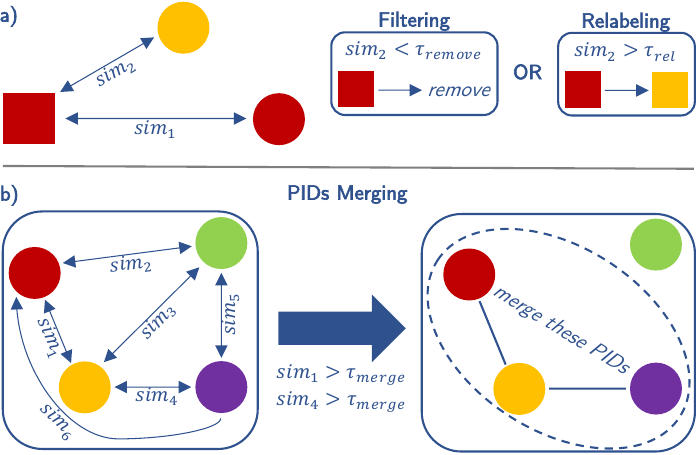}
    \caption{Illustration of the Relabeling Module stages. (a)~Filtering \& Relabeling: the module assesses the similarity between feature vectors and centroids, removing images with low similarity and updating pseudo-labels if a better match is found. (b) PIDs Merging: constructs a graph based on centroid similarity where centroids with strong connections are merged, consolidating PIDs that likely belong to the same individual. In this illustration, circles represent centroids, and squares denote feature vectors for images. Matching colors indicate similarity in PIDs.}
    \label{fig:relabeling-module}
\end{figure}

Pseudo-labels in single-camera data are obtained through person tracking and are inherently noisy. Some images cannot be confidently assigned to a specific PID, others might be assigned to the PID of another person. In some cases, a person gets assigned multiple PIDs. We address three types of errors with filtering, relabeling, and merging PIDs, respectively (Fig.~\ref{fig:relabeling-module}).

To mitigate label noise, DynaMix introduces the Relabeling Module, which refines pseudo-labels dynamically at the beginning of each training epoch. Unlike prior methods, this procedure is performed on-the-fly and scales efficiently to large-scale single-camera data with millions of images and hundreds of thousands of identities.

To keep pseudo-labels refinement efficient,  at each epoch the module randomly selects $K$ images $x_i$ per pseudo-label $y_i \in \mathcal{Y}_s$, and feature vectors $f_i$ are obtained using the momentum encoder from the previous epoch.

\subsubsection{Filtering \& Relabeling}
\label{sec:relabeling-module-filtering-and-relabeling}

The Relabeling Module refines pseudo-labels by comparing feature vectors $f_i$ with centroids $\mu_k \in \mathcal{M}_s$. Let $y_i$ be the current pseudo-label for $x_i$, and $y_j$ correspond to the centroid $\mu_{y_j} = \arg\max_k sim(f_i, \mu_k))$, where $y_j \neq y_i$. Based on the maximum similarity score, we apply:
\begin{itemize}\vspace{-6pt}
    \setlength\itemsep{-0.1em}
   \item \textbf{Filtering:} if $sim(f_i, \mu_{y_j}) < \tau_{remove}$, $x_i$ is deemed insufficiently similar to any centroid and is removed from the data for this epoch, eliminating noisy images.
   \item \textbf{Relabeling:} if $sim(f_i, \mu_{y_j}) > \tau_{rel} > \tau_{remove}$, the pseudo-label for $x_i$ is updated from $y_i$ to $y_j$, indicating a closer match to another PID’s centroid.
\end{itemize}

\subsubsection{PIDs Merging}

PIDs that likely belong to the same individual are merged by leveraging pairwise similarity between centroids $\mu_k \in \mathcal{M}_s$. The process is as follows:
\begin{enumerate}\vspace{-6pt}
    \setlength\itemsep{-0.1em}
    \item Compute the pairwise similarity matrix $S$: $S_{ij} = sim(\mu_i, \mu_j)$.
    \item Treat $S$ as an adjacency matrix of a graph, where nodes are centroids. Retain only edges with $S_{ij} \geq \tau_{merge}$, setting other entries to $0$ to create a binary adjacency matrix.
    \item Identify connected components in the resulting graph. Centroids within the same component are assumed to belong to the same individual.
    \item Merge all PIDs within each connected component, unifying fragmented identities.
\end{enumerate}\vspace{-6pt}

\noindent This procedure is repeated at the beginning of each epoch, dynamically refining PIDs.

\subsection{The Efficient Centroids Module}
\label{sec:efficient-centroids-module}

The Efficient Centroids Module maintains centroids for all identity pseudo-labels in $\mathcal{D}_s$, using the refined pseudo-labels produced by the Relabeling Module. These centroids are stored in the Centroids Memory $\mathcal{M}_s$ and are shared with other modules throughout the training process. As the momentum encoder is continuously updated, centroids must also be updated every epoch.

Since single-camera data includes a large number of PIDs, recomputing all centroids from scratch would be computationally expensive. To address this, DynaMix employs an efficient dynamic update strategy based on an exponential moving average. At the start of each epoch, feature vectors $f_i$ are computed for the same randomly selected images in the Relabeling Module (Sec.~\ref{sec:relabeling-module}), and centroids are updated as:
\begin{equation}
      \mu_k \leftarrow \alpha \mu_k + (1 - \alpha) \frac{1}{|I_k|} \sum_{i \in I_k} f_i,
    \label{eq:efficient-centroids-momentum}
\end{equation}
where $\alpha$ is a smoothing factor, and $I_k$ is the set of indices for images with pseudo-label $k$. 

The dynamic updating enables efficient and scalable centroid computation (see Sec.~\ref{sec:efficient-centroids-module-efficiency} for efficiency analysis). Besides, by leveraging the refined pseudo-labels from the Relabeling Module, it provides more accurate and stable identity representations.

\begin{table*}[t]
    \centering
    \footnotesize
    \begin{tabular*}{1.0\textwidth}{@{\hspace{9pt}}lcccccccc}
        \toprule
        \multirow{2}{*}{Dataset} & \multirow{2}{*}{Type} & \multirow{2}{*}{Environment} & \multirow{2}{*}{Labels} & \multicolumn{3}{c}{\#Images} & \multirow{2}{*}{\#PIDs} & \multirow{2}{*}{\#Scenes}\\
        \cmidrule{5-7}
        & & & & Train & Query & Gallery & &\rule[-1.0ex]{0pt}{0pt}\\
        \midrule
        CUHK03-NP \cite{li2014deepreid} & \multirow{6}{*}{multi-camera} & campus & manual & 7,365 & 1,400 & 5,332 & 1,467 & 2\rule{0pt}{2.3ex}\\
        Market-1501 \cite{zheng2015scalable} & & campus & manual & 12,936 & 3,368 & 15,913 & 1,501 & 6\rule{0pt}{2.3ex}\\
        MSMT17 \cite{wei2018person} & & campus & manual & 32,621 & 11,659 & 82,161 & 4,101 & 15\rule{0pt}{2.3ex}\\
        MSMT17-merged \cite{wei2018person} & & campus & manual & 126,441 & --- & --- & 4,101 & 15\rule{0pt}{2.3ex}\\
        DukeMTMC-reID \cite{ristani2016performance} & & campus & manual & 16,522 & 2,228 & 17,661 & 1,812 & 8\rule{0pt}{2.3ex}\\
        RandPerson  (part) \cite{wang2020surpassing} & & synthetic & synthetic & 132,145 & --- & --- & 8,000 & 19\rule{0pt}{2.1ex}\\
        \midrule
        LUPerson-NL \cite{fu2022large} & single-camera & vary & pseudo & $>$10M & --- & --- & 434K & 21,697\rule{0pt}{2.1ex}\\
        \bottomrule
    \end{tabular*}
    \caption{Statistics of datasets used in our experiments. The single-camera LUPerson-NL is orders of magnitude larger than all multi-camera datasets in the number images, identities, as well as environments variety.}
    \label{tab:datasets}
\end{table*}

\subsection{The Data Sampling Module}
\label{sec:data-sampling-module}

Generalization in Re-ID requires mini-batches that are both complex and diverse. In DynaMix, this is especially important due to the heterogeneous nature of the data --- multi-camera and single-camera datasets. Effective mini-batch composition must account for both intra-batch diversity and the relative complexity of each data type within the mini-batch.

To this end, we propose the Data Sampling Module that constructs mixed data mini-batches with balanced learning complexity and diversity:
\begin{enumerate}\vspace{-6pt}
    \setlength\itemsep{-0.1em}
    \item Randomly select $N_P$ PIDs $y_m^i \in \mathcal{Y}_m$ and compute centroids $\widetilde{\mu}_{y_m^i}$.
    \item Construct a similarity matrix $S \in \mathbb{R}^{N_P \times M_s}$, where~$S_{ij} = sim(\widetilde{\mu}_{y_m^i}, \mu_{y_s^j})$ represents similarity between multi-camera centroids and each single-camera centroid $\mu_{y_s^j} \in \mathcal{M}_s$, maintained by the Efficient Centroids Module (Sec.~\ref{sec:efficient-centroids-module}).
    \item Use the Hungarian algorithm \cite{kuhn1955hungarian} to find a globally optimal one-to-one assignment between multi-camera PIDs $y_m^i$ and single-camera PIDs $y_s^j$ based on the similarity matrix $S$. Retain only the matched pairs whose similarity is close to the median across all combinations, avoiding overly easy or overly hard pairs to maintain a balance between learning complexity and diversity (Sec.~\ref{sec:similarity-score-selection-method}).
    \item Maintain a FIFO-queue of recently used single-camera PIDs. A single-camera PID $y_s^j$ is excluded from the current assignment step. The queue is updated after each mini-batch to promote the diversity of consecutive single-camera data samples (Sec.~\ref{sec:queue-size-importance}).
   \item For each selected pair $(y_m^i, y_s^j)$, sample $N_K$ images from both data sources. For multi-camera data, images are selected from different cameras. The final mixed data mini-batch contains $2 \times N_P \times N_K$ images.
\end{enumerate}\vspace{-6pt}
Since data sampling in DynaMix aims to pair diverse single-camera and multi-camera samples within each mini-batch, this process requires finding a global optimum rather than relying on a greedy heuristic. Therefore, DynaMix uses the Hungarian algorithm, which guarantees such an optimal assignment. Importantly, due to the relatively small mini-batch size used during training~(Sec.~\ref{sec:implementation-details}), the computational overhead introduced by the algorithm remains negligible.

\section{Experiments}
\label{sec:experiments}

\subsection{Datasets and Evaluation Metrics}

\vspace{2pt}\noindent\textbf{Multi-camera datasets.} We evaluate the proposed DynaMix method on well-known multi-camera datasets: CUHK03-NP \cite{li2014deepreid}, Market-1501~\cite{zheng2015scalable}, MSMT17 \cite{wei2018person} with its modification MSMT17-merged, which combines training and testing parts, DukeMTMC-reID \cite{ristani2016performance}, and a subset of RandPerson \cite{wang2020surpassing} (Tab.~\ref{tab:datasets}). DukeMTMC-reID was withdrawn over ethical concerns, but is included here due to its widespread use in multi-source cross-dataset evaluations. 

\vspace{2pt}\noindent\textbf{Single-camera dataset.} We use LUPerson-NL~\cite{fu2022large} as a pseudo-labeled single-camera dataset. This dataset was created by processing YouTube videos and automatically annotated via tracking. As seen in Tab.~\ref{tab:datasets}, LUPerson-NL is significantly larger than traditional multi-camera Re-ID datasets and offers a much broader diversity.

\vspace{2pt}\noindent\textbf{Metrics.} In our experiments, we use Cumulative Matching Characteristics ($Rank_1$) and mean Average Precision ($mAP$) as evaluation metrics.

\subsection{Implementation Details}
\label{sec:implementation-details}

Both encoder and momentum encoder are ViT-base \cite{dosovitskiy2020image} pre-trained on the single-camera LUPerson \cite{fu2021unsupervised} dataset in a self-supervised manner. We employ SGD with a learning rate of $10^{-3}$ and a weight decay rate of $10^{-4}$. A warm-up scheme is applied during the first 10 epochs. DynaMix is trained for 100 epochs, with $N_P = 8$ and $N_K = 4$, resulting in a mini-batch size of 64 images. Each epoch consists of $400$ iterations, as in ReMix. The momentum coefficient $\lambda$ in Eq.~\ref{eq:method-momentum} is set to $0.999$. All images are resized to $256 \times 128$ and augmented with random cropping, horizontal flipping, Gaussian blurring, and random grayscale transformations. DynaMix is trained on $2$ Nvidia RTX $3090$ GPUs under PyTorch framework \cite{paszke2019pytorch}.

\subsection{Comparison with State-of-the-Art Methods}
\label{sec:comparison}

\begin{table*}[!t]
    \centering
    \begin{subtable}[b]{\linewidth}
        \centering
        \footnotesize
        \begin{tabular}{@{\hspace{11pt}}l|c|c|cc|cc|cc}
            \toprule
            \multirow{2}{*}{Method} & \multirow{2}{*}{Reference} & \multirow{2}{*}{Training Dataset} & \multicolumn{2}{c|}{CUHK03-NP} & \multicolumn{2}{c|}{Market-1501} & \multicolumn{2}{c}{MSMT17}\\
            & & & $Rank_1$ & $mAP$ & $Rank_1$ & $mAP$ & $Rank_1$ & $mAP$ \rule[-0.5ex]{0pt}{0pt}\\
            \midrule
            TransMatcher \cite{liao2021transmatcher} & NeurIPS21 & \multirow{7}{*}{Market-1501} & $22.2$ & $21.4$ & --- & --- & $47.3$ & $18.4$ \rule{0pt}{2.3ex}\\
            QAConv-GS \cite{liao2022graph} & CVPR22 & & $19.1$ & $18.1$ & $91.6$ & $75.5$ & $45.9$ & $17.2$ \rule{0pt}{2.3ex}\\
            PAT \cite{ni2023part} & ICCV23 & & $25.4$ & $26.0$ & $92.4$ & $81.5$ & $42.8$ & $18.2$ \rule{0pt}{2.3ex}\\
            LDU \cite{peng2024invariance} & TIM24 & & $18.5$ & $18.2$ & --- & --- & $35.7$ & $13.5$ \rule{0pt}{2.3ex}\\
            DCAC \cite{li2025unleashing} & Sensors25 & & $33.2$ & $32.5$ & $94.9$ & $86.8$ & $52.1$ & $23.4$ \rule{0pt}{2.3ex}\\
            ReMix \cite{mamedov2025remix} & WACV25 & & --- & --- & $96.2$ & $89.8$ & --- & --- \rule{0pt}{2.3ex}\\
            DynaMix & Ours & & $\mathbf{51.0}$ & $\mathbf{52.2}$ & $\mathbf{97.4}$ & $\mathbf{93.8}$ & $\mathbf{65.5}$ & $\mathbf{36.7}$ \rule{0pt}{2.1ex}\\
            \midrule
            TransMatcher \cite{liao2021transmatcher} & NeurIPS21 & \multirow{7}{*}{MSMT17} & $23.7$ & $22.5$ & $80.1$ & $52.0$ & --- & --- \rule{0pt}{2.3ex}\\
            QAConv-GS \cite{liao2022graph} & CVPR22 & & $20.9$ & $20.6$ & $79.1$ & $49.5$ & $79.2$ & $50.9$ \rule{0pt}{2.3ex}\\
            PAT \cite{ni2023part} & ICCV23 & & $24.2$ & $25.1$ & $72.2$ & $47.3$ & $75.9$ & $52.0$ \rule{0pt}{2.3ex}\\
            LDU \cite{peng2024invariance} & TIM24 & & $21.3$ & $21.3$ & $74.6$ & $44.8$ & --- & --- \rule{0pt}{2.3ex}\\
            DCAC \cite{li2025unleashing} & Sensors25 & & $34.4$ & $34.1$ & $77.9$ & $52.1$ & $88.3$ & $70.1$ \rule{0pt}{2.3ex}\\
            ReMix \cite{mamedov2025remix} & WACV25 & & $27.3$ & $27.4$ & $78.2$ & $52.4$ & $84.8$ & $63.9$ \rule{0pt}{2.3ex}\\
            DynaMix & Ours & & $\mathbf{48.9}$ & $\mathbf{49.6}$ & $\mathbf{91.1}$ & $\mathbf{77.7}$ & $\mathbf{90.4}$ & $\mathbf{76.3}$ \rule{0pt}{2.1ex}\\
            \midrule
            QAConv \cite{liao2020interpretable} & ECCV20 & \multirow{7}{*}{MSMT17-merged} & $25.3$ & $22.6$ & $72.6$ & $43.1$ & \cellcolor{gray!20} & \cellcolor{gray!20} \rule{0pt}{2.3ex}\\
            TransMatcher \cite{liao2021transmatcher} & NeurIPS21 & & $31.9$ & $30.7$ & $82.6$ & $58.4$ & \cellcolor{gray!20} & \cellcolor{gray!20} \rule{0pt}{2.3ex}\\
            QAConv-GS \cite{liao2022graph} & CVPR22 & & $27.6$ & $28.0$ & $82.4$ & $56.9$ & \cellcolor{gray!20} & \cellcolor{gray!20} \rule{0pt}{2.3ex}\\
            PAT \cite{ni2023part} & ICCV23 & & $27.4$ & $28.7$ & $72.8$ & $48.6$ & \cellcolor{gray!20} & \cellcolor{gray!20} \rule{0pt}{2.3ex}\\
            CLIP-DFGS \cite{zhao2024clip} & TOMM24 & & $37.1$ & $25.7$ & $80.9$ & $55.2$ & \cellcolor{gray!20} & \cellcolor{gray!20} \rule{0pt}{2.3ex}\\
            ReMix \cite{mamedov2025remix} & WACV25 & & $37.7$ & $37.2$ & $84.0$ & $61.0$ & \cellcolor{gray!20} & \cellcolor{gray!20} \rule{0pt}{2.3ex}\\
            DynaMix & Ours & & $\mathbf{60.9}$ & $\mathbf{61.0}$ & $\mathbf{92.6}$ & $\mathbf{80.8}$ & \cellcolor{gray!20} & \cellcolor{gray!20} \rule{0pt}{2.1ex}\\
            \midrule
            TransMatcher \cite{liao2021transmatcher} & NeurIPS21 & \multirow{5}{*}{RandPerson} & $17.1$ & $16.0$ & $77.3$ & $49.1$ & $48.3$ & $17.7$ \rule{0pt}{2.3ex}\\
            QAConv-GS \cite{liao2022graph} & CVPR22 & & $18.4$ & $16.1$ & $76.7$ & $46.7$ & $45.1$ & $15.5$ \rule{0pt}{2.3ex}\\
            PAT \cite{ni2023part} & ICCV23 & & $20.2$ & $20.1$ & $73.7$ & $46.9$ & $45.5$ & $19.4$ \rule{0pt}{2.3ex}\\
            ReMix \cite{mamedov2025remix} & WACV25 & & $19.3$ & $18.4$ & $72.7$ & $45.4$ & --- & --- \rule{0pt}{2.3ex}\\
            DynaMix & Ours & & $\mathbf{36.9}$ & $\mathbf{39.3}$ & $\mathbf{89.3}$ & $\mathbf{72.4}$ & $\mathbf{67.1}$ & $\mathbf{36.5}$ \rule{0pt}{2.1ex}\\
            \bottomrule
        \end{tabular}
        \caption{The cross-dataset scenario. In this table, gray cells indicate experiments that are not applicable.}
        \label{tab:comparison-single}
    \end{subtable}
    
    \vspace{0.15cm}

    \begin{subtable}[b]{\linewidth}
        \centering
        \footnotesize
        \begin{tabularx}{0.91\linewidth}{@{\hspace{7pt}}l|c|cc|cc|cc|cc}
            \toprule
            \multirow{2}{*}{Method} & \multirow{2}{*}{Reference} & \multicolumn{2}{c|}{M+D+MS $\rightarrow$ C3} & \multicolumn{2}{c|}{D+C3+MS $\rightarrow$ M} & \multicolumn{2}{c|}{M+C3+MS $\rightarrow$ D} & \multicolumn{2}{c}{M+D+C3 $\rightarrow$ MS}\\
            & & $Rank_1$ & $mAP$ & $Rank_1$ & $mAP$ & $Rank_1$ & $mAP$ & $Rank_1$ & $mAP$\rule{0pt}{2.3ex}\rule[-0.5ex]{0pt}{0pt}\\
            \midrule
            MetaBIN \cite{choi2021meta} & CVPR21 & $38.1$ & $37.5$ & $83.2$ & $61.2$ & $71.3$ & $54.9$ & $40.8$ & $17.0$ \rule{0pt}{2.3ex}\\
            MixNorm \cite{qi2022novel} & TMM22 & $29.6$ & $29.0$ & $78.9$ & $51.4$ & $70.8$ & $49.9$ & $47.2$ & $19.4$ \rule{0pt}{2.3ex}\\
            META \cite{xu2021meta} & ECCV22 & $46.0$ & $45.9$ & $85.3$ & $65.7$ & $76.9$ & $59.9$ & $49.3$ & $22.5$ \rule{0pt}{2.3ex}\\
            IL \cite{tan2023style} & TMM23 & $40.9$ & $38.3$ & $86.2$ & $65.8$ & $75.4$ & $57.1$ & $45.7$ & $20.2$ \rule{0pt}{2.3ex}\\
            ISR \cite{dou2023identity} & ICCV23 & $36.6$ & $37.8$ & $87.0$ & $70.5$ & --- & --- & $56.4$ & $30.3$ \rule{0pt}{2.3ex}\\
            UDSX \cite{ang2024unified} & Neuro24 & $38.9$ & $37.2$ & $85.7$ & $60.4$ & $74.7$ & $55.8$ & $47.6$ & $20.2$ \rule{0pt}{2.3ex}\\
            BAU \cite{cho2024generalizable} & NeurIPS24 & $40.4$ & $40.4$ & $81.9$ & $56.7$ & $74.4$ & $57.4$ & $45.3$ & $19.4$ \rule{0pt}{2.3ex}\\
            DCAC \cite{li2025unleashing} & Sensors25 & $43.6$ & $42.5$ & $80.0$ & $56.7$ & $75.4$ & $58.9$ & $56.7$ & $27.5$ \rule{0pt}{2.3ex}\\
            ReMix \cite{mamedov2025remix} & WACV25 & $47.6$ & $46.5$ & $87.8$ & $70.5$ & $79.0$ & $63.3$ & --- & --- \rule{0pt}{2.3ex}\\
            DynaMix & Ours & $\mathbf{66.6}$ & $\mathbf{67.1}$ & $\mathbf{93.3}$ & $\mathbf{83.3}$ & $\mathbf{84.6}$ & $\mathbf{74.9}$ & $\mathbf{71.0}$ & $\mathbf{43.2}$ \rule{0pt}{2.1ex}\\
            \bottomrule
        \end{tabularx}
        \caption{The multi-source cross-dataset scenario. In this table, C3 is CUHK03-NP, M is Market-1501, D is DukeMTMC-reID, and MS is MSMT17.}
        \label{tab:comparison-multi}
    \end{subtable}
    \caption{Comparison of DynaMix with other state-of-the-art methods in both cross-dataset and multi-source cross-dataset scenarios.}
    \label{tab:comparison}
\end{table*}

To assess the generalization ability of DynaMix and existing approaches, we use two conventional evaluation protocols: cross-dataset and multi-source cross-dataset. In the cross-dataset protocol, a model is trained on one multi-camera dataset and tested on another multi-camera dataset. In the multi-source cross-dataset protocol, several multi-camera datasets are used for training. Additionally, we report the results of the standard Re-ID evaluation, in which models are trained and tested on different splits of the same dataset.

\vspace{2pt}\noindent\textbf{The cross-dataset scenario.} According to Tab.~\ref{tab:comparison-single}, DynaMix consistently outperforms state-of-the-art methods in all benchmarks. In RandPerson $\rightarrow$ CUHK03-NP, the improvement is almost twofold. DynaMix achieves significant gains over transformer-based TransMatcher \cite{liao2021transmatcher} and PAT \cite{ni2023part}, showing that its superior generalization comes from effective training on mixed data rather than a more powerful encoder. Furthermore, DynaMix is superior to QAConv \cite{liao2020interpretable}, QAConv-GS \cite{liao2022graph}, and TransMatcher \cite{liao2021transmatcher} that use larger input image sizes.

\vspace{2pt}\noindent\textbf{The multi-source cross-dataset scenario.} When trained on several multi-camera datasets, DynaMix exhibits outstanding performance across datasets (Tab.~\ref{tab:comparison-multi}). This study demonstrates its ability to effectively utilize single-camera data along with multi-camera data from diverse domains.

\vspace{2pt}\noindent\textbf{The standard Re-ID scenario.} Although we primarily target generalizable person Re-ID, we also compare DynaMix with other methods in the standard Re-ID same-dataset scenario. As can be seen from Tab.~\ref{tab:comparison-single}, DynaMix outperforms competing approaches when trained and tested within a single dataset. This result highlights the flexibility and strong performance of DynaMix in various scenarios, including those for which it is not originally designed.

\vspace{2pt}\noindent\textbf{Comparison conclusion.} Only ReMix and DynaMix are trained using additional single-camera data, which is central to the design of both methods. All other approaches rely solely on labeled multi-camera datasets, as they are not created to handle heterogeneous training data. Therefore, adapting them to use single-camera data would lead to an unfair degradation of their performance. For this reason, their results are reported under the official training settings. Our experiments show that ReMix is less effective at leveraging single-camera data compared to DynaMix, which further highlights the advantage of the proposed method.

\subsection{Ablation Study}
\label{sec:ablation}

\begin{figure*}[th]
    \centering
    \includegraphics[width=0.97\linewidth]{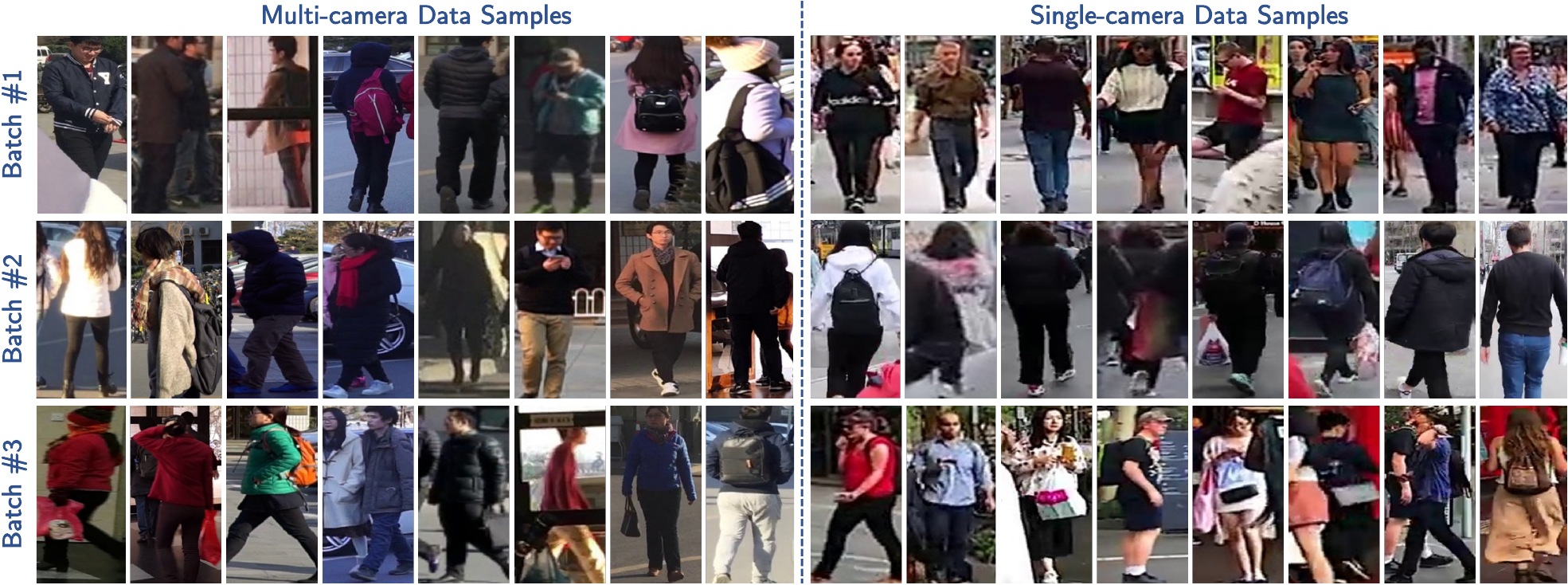}
    \caption{Examples of mini-batches composed by the Data Sampling Module. Single-camera samples are selected to visually match multi-camera data within mini-batch, while maintaining stylistic diversity across batches to balance learning complexity and data variety.}
    \label{fig:data-sampling-examples}
\end{figure*}

\begin{table}[t]
    \centering
    \footnotesize
    \begin{tabular*}{0.39\textwidth}{@{\hspace{5pt}}cccc|cc}
        \toprule
        ViT & ECM & DSM & RM & $Rank_1$ & $mAP$ \rule{0pt}{2.3ex}\rule[-0.9ex]{0pt}{0pt}\\
        \midrule
        & & & & $83.3$ & $60.1$ \rule{0pt}{2.3ex}\\
        \ding{51} & & & & $84.5$ & $63.9$ \rule{0pt}{2.3ex}\\
        \ding{51} & \ding{51} & & & $84.9$ & $65.4$ \rule{0pt}{2.3ex}\\
        \ding{51} & \ding{51} & \ding{51} & & $86.8$ & $71.7$ \rule{0pt}{2.3ex}\\
        \ding{51} & \ding{51} & \ding{51} & \ding{51} & $\mathbf{92.6}$ & $\mathbf{80.8}$ \rule{0pt}{2.1ex}\\
        \bottomrule
    \end{tabular*}
    \caption{Step-by-step ablation study of DynaMix components. Here, ViT is Vision Transformer, ECM is the Efficient Centroids Module, DSM is the Data Sampling Module, and RM is the Relabeling Module. The model is trained on MSMT17-merged and tested on Market-1501. The first row corresponds to a baseline (ReMix).}
    \label{tab:ablation-overall}
\end{table}

DynaMix introduces several novel components that enhance Re-ID performance. We conduct an ablation study by incrementally adding each component and evaluating its contribution. According to Tab.~\ref{tab:ablation-overall}, each module improves accuracy, and their combination yields the best overall performance.

To better highlight the effectiveness of the proposed components, we use ReMix as the baseline, adapted to the DynaMix training setup. Specifically, the baseline employs the pseudo-labeled LUPerson-NL dataset (instead of LUPerson used in the original ReMix) to match our data pipeline. This adjustment ensures a fair and consistent comparison by evaluating both methods under identical data conditions.

\subsubsection{Vision Transformer}
\label{sec:vit}

Since the baseline ReMix employs a ResNet50-IBN backbone \cite{he2016deep, pan2018two}, while DynaMix is built upon a ViT encoder, we include an explicit backbone replacement step in our ablation study (second row in Tab.~\ref{tab:ablation-overall}). This ensures a fair and transparent comparison by isolating the effect of the architectural change from the contributions of our newly introduced modules.

Our experiments show that switching to a ViT backbone improves the generalization ability. An additional practical benefit is that ViT-base produces $768$-dimensional embeddings instead of the $2048$-dimensional embeddings from ResNet50-IBN, resulting in over $2.5\times$ faster image retrieval --- a particularly valuable advantage for large-scale real-world deployments.

\subsubsection{The Efficient Centroids Module}

Since both the Data Sampling Module and the Relabeling Module rely on identity centroids computed from large-scale single-camera data, it is essential to keep them accurate and up-to-date throughout training. To this end, we dynamically update centroids at the start of each epoch. As shown in Tab.~\ref{tab:ablation-overall}, the dynamic centroid update in the Efficient Centroids Module achieves comparable performance to its naive counterpart, which uses all images. Moreover, it improves cross-dataset Re-ID accuracy. We attribute this to reducing noise when using refined pseudo-labels obtained by the Relabeling Module.

\subsubsection{The Data Sampling Module}

In DynaMix, the Data Sampling Module plays a key role in composing effective mini-batches from heterogeneous data (Sec.~\ref{sec:data-sampling-module}). According to Tab.~\ref{tab:ablation-overall}, replacing naive random sampling with our module significantly improves generalizable Re-ID performance. We attribute this to increased learning complexity and higher intra-batch diversity. As demonstrated in Fig.~\ref{fig:data-sampling-examples}, the Data Sampling Module balances learning complexity and data variety.

\subsubsection{The Relabeling Module}

The largest accuracy gain is achieved through on-the-fly relabeling of single-camera data (Tab.~\ref{tab:ablation-overall}). Since ground-truth labels are unavailable for single-camera data, the most reliable way to assess relabeling quality is through its impact on downstream Re-ID accuracy. The substantial improvement observed supports the effectiveness of our Relabeling Module. In addition, Tab.~\ref{tab:relabeling-stats} reports statistics on the number of images and pseudo-labels before and after relabeling. Notably, this module is closely connected to~other DynaMix components and benefits from their synergy.

\begin{table}[t]
    \centering
    \footnotesize
    \begin{tabular*}{0.4\textwidth}{@{\hspace{8pt}}l|ccc}
        \toprule
        Statistics & Before & After & Delta \rule{0pt}{2.3ex}\rule[-0.9ex]{0pt}{0pt}\\
        \midrule
        \#Images & 10,683,716 & 9,889,173 & $-$7.4\% \rule{0pt}{2.3ex}\\
        \#PIDs & 433,997 & 364,490 & $-$16.0\% \rule{0pt}{2.1ex}\\
        \bottomrule
    \end{tabular*}
    \caption{Impact of the Relabeling Module on the single-camera LUPerson-NL dataset. The table shows the number of images and PIDs before and after relabeling.}
    \label{tab:relabeling-stats}
\end{table}

\subsection{Parameter Analysis}
\label{sec:parameter}

\subsubsection{Efficiency Analysis of the Efficient Centroids Module}
\label{sec:efficient-centroids-module-efficiency}

\begin{table}[t]
    \centering
    \footnotesize
    \begin{tabular*}{0.2\textwidth}{@{\hspace{7pt}\extracolsep{\fill}}c|*{2}{c}@{\hspace{4pt}}}
        \toprule
        $K$ & $Rank_1$ & $mAP$ \rule{0pt}{2.3ex}\rule[-0.9ex]{0pt}{0pt}\\
        \midrule
        $2$ & $84.8$ & $65.2$ \rule{0pt}{2.3ex}\\
        $4$ & $84.9$ & $65.4$ \rule{0pt}{2.3ex}\\
        $8$ & $\mathbf{85.2}$ & $\mathbf{65.9}$ \rule{0pt}{2.1ex}\\
        \bottomrule
    \end{tabular*}
    \caption{Analysis of the number of randomly selected images per person identity ($K$) in the Efficient Centroids Module. The model is trained on MSMT17-merged and tested on Market-1501.}
    \label{tab:k-param}
\end{table}

\begin{table}[t]
    \centering
    \footnotesize
    \begin{tabular*}{0.27\textwidth}{@{\hspace{6pt}\extracolsep{\fill}}c|cc}
        \toprule
        $K$ & Runtime & Memory \rule{0pt}{2.3ex}\rule[-0.9ex]{0pt}{0pt}\\
        \midrule
        $2$ & $8$ min  & $2.5$ GB \rule{0pt}{2.3ex}\\
        $4$ & $15$ min & $4.9$ GB \rule{0pt}{2.3ex}\\
        $8$ & $30$ min & $9.8$ GB \rule{0pt}{2.1ex}\\
        \midrule
        Naive & $90$ min & $29.5$ GB \rule{0pt}{2.1ex}\\
        \bottomrule
    \end{tabular*}
    \caption{Analysis of runtime and memory consumption for different values of $K$ in the Efficient Centroids Module.}
    \label{tab:runtime-memory}
\end{table}

Tab.~\ref{tab:k-param} analyzes the effect of the number of randomly selected images per person identity ($K$) on performance. As expected, larger values of $K$ improve accuracy, but also increase computational cost. To balance performance and efficiency, we select $K=4$, enabling robust centroid updates that incorporate refined pseudo-labels from the Relabeling Module effectively in each epoch.

This strategy is crucial for large-scale single-camera data, where recomputing all centroids from scratch every epoch would be computationally expensive. For instance, in the LUPerson-NL dataset, full centroid recomputation requires $>$10M embeddings per epoch. In contrast, DynaMix computes only 4 $\times$ \#PIDs $=$ 4 $\times$ 434K $=$ 1.7M embeddings, which is $\approx$6x fewer. 

As shown in Tab.~\ref{tab:runtime-memory}, the runtime of centroid updates grows approximately linearly with $K$. For $K=2,4,8$, the update step takes $8$, $15$, and $30$ minutes per epoch, respectively. In contrast, the naive approach --- recomputing centroids from all $>$10M images --- requires around $90$ minutes per epoch under the same hardware configuration~(Sec.~\ref{sec:implementation-details}). Memory usage follows the same pattern. Since ViT-base produces $768$-dimensional embeddings (Sec.~\ref{sec:vit}), centroid updates require $4.9$ GB for $K=4$. The naive method consumes $\approx$$29.5$ GB, consistent with its significantly larger computational load.

Together, Tab.~\ref{tab:k-param} and Tab.~\ref{tab:runtime-memory} demonstrate that $K=4$ offers the best trade-off between accuracy and efficiency. Moreover, as shown in the ablation study (Tab.~\ref{tab:ablation-overall}), the Efficient Centroids Module not only reduces time and memory costs but also improves Re-ID accuracy due to its synergy with the Relabeling Module --- unlike the naive alternative.

\subsubsection{Analysis of the Smoothing Factor}
\label{sec:smothing-factor-analysis}

In Tab.~\ref{tab:alpha-param}, we analyze the effect of the smoothing factor ($\alpha$) used for updating centroids in the Efficient Centroids Module. This parameter controls the balance between previously accumulated centroids and newly computed feature vectors at each epoch. Lower values of $\alpha$ lead to faster adaptation but may introduce instability, while higher values provide smoother updates but reduce responsiveness to recent changes. We set $\alpha = 0.3$ as it offers the best trade-off between stability and adaptability, resulting in improved generalization in person Re-ID.

\begin{table}[t]
    \centering
    \footnotesize
    \begin{tabular*}{0.25\textwidth}{@{\hspace{7pt}\extracolsep{\fill}}c|*{2}{c}@{\hspace{3pt}}}
        \toprule
        $\alpha$ & $Rank_1$ & $mAP$ \rule{0pt}{2.3ex}\rule[-0.9ex]{0pt}{0pt}\\
        \midrule
        $0.2$ & $84.2$ & $65.0$ \rule{0pt}{2.3ex}\\
        $0.3$ & $\mathbf{84.9}$ & $\mathbf{65.4}$ \rule{0pt}{2.3ex}\\
        $0.4$ & $84.8$ & $65.2$ \rule{0pt}{2.3ex}\\
        $0.5$ & $84.7$ & $64.9$ \rule{0pt}{2.3ex}\\
        $0.7$ & $83.9$ & $64.6$ \rule{0pt}{2.1ex}\\
        \bottomrule
    \end{tabular*}
    \caption{Analysis of the smoothing factor for updating centroids ($\alpha$) in the Efficient Centroids Module. The model is trained on MSMT17-merged and tested on Market-1501.}
    \label{tab:alpha-param}
\end{table}

\subsubsection{Similarity Score Selection Method}
\label{sec:similarity-score-selection-method}

Similarity score selection is crucial for balancing learning complexity and intra-batch diversity in the Data Sampling Module. In Tab.~\ref{tab:sim-score-select-method}, we analyze the impact of different similarity score selection methods on generalizable Re-ID performance.

\textit{Hard} similarity selection pairs each multi-camera PID with the most similar single-camera PID. Although such pairs increase learning difficulty (since the ID spaces of multi-camera and single-camera data do not overlap), this strategy has two major drawbacks. First, hard matching often produces mini-batches containing stylistically similar individuals (e.g., people wearing dark clothing), which substantially reduces appearance diversity and generalization. Second, these highly similar images may correspond to fragmented pseudo-labels of the same person. As a result, the model may receive incorrect supervision, which leads to overfitting to pseudo-label noise and ultimately harms generalization.

\textit{Soft} similarity selection, in contrast, prioritizes the least similar pairs and thus increases visual diversity. However, this strategy makes the learning task overly easy: the multi-camera and single-camera images in a mini-batch become semantically unrelated, resulting in uninformative gradients. Consequently, the model is not encouraged to focus on subtle person-specific differences --- an essential capability for Re-ID --- which ultimately limits its ability to generalize.

Selecting \textit{median} similarity scores provides a balanced and informative supervision signal. Median pairs are neither dominated by pseudo-label noise~(as in hard pairs) nor trivially unrelated~(as in soft pairs). Instead, they retain sufficient stylistic diversity while still being semantically meaningful, allowing the model to learn discriminative features. This explains why median similarity sampling leads to the best performance in the cross-dataset scenario.

\begin{table}[t]
    \centering
    \footnotesize
    \begin{tabular*}{0.27\textwidth}{@{\hspace{8pt}\extracolsep{\fill}}l|*{2}{c}}
        \toprule
        $Method$ & $Rank_1$ & $mAP$ \rule{0pt}{2.3ex}\rule[-0.9ex]{0pt}{0pt}\\
        \midrule
        $random$ & $85.6$ & $69.9$ \rule{0pt}{2.3ex}\\
        $hard$ & $84.4$ & $64.9$ \rule{0pt}{2.3ex}\\
        $soft$ & $84.7$ & $68.1$ \rule{0pt}{2.3ex}\\
        $mean$ & $86.7$ & $70.8$ \rule{0pt}{2.3ex}\\
        $median$ & $\mathbf{86.8}$ & $\mathbf{71.7}$ \rule{0pt}{2.1ex}\\
        \bottomrule
    \end{tabular*}
    \caption{Analysis of the similarity score selection method in the Data Sampling Module. The model is trained on MSMT17-merged and tested on Market-1501.}
    \label{tab:sim-score-select-method}
\end{table}

\subsubsection{Analysis of the Importance of Queue Size}
\label{sec:queue-size-importance}

The queue size in the Data Sampling Module plays a critical role in maintaining intra-batch diversity by preventing immediate reuse of single-camera PIDs across consecutive mini-batches. As shown in Tab.~\ref{tab:queue-length-param}, increasing the queue size initially improves generalization by promoting diversity. However, excessively large queues lead to performance degradation due to reduced sampling flexibility and less relevant identity pools. Based on this analysis, the size of 30 epochs is chosen as optimal to support robust generalization.

\begin{table}[t]
    \centering
    \footnotesize
    \begin{tabular*}{0.25\textwidth}{@{\hspace{6pt}\extracolsep{\fill}}c|*{2}{c}@{\hspace{4pt}}}
        \toprule
        $Size$ & $Rank_1$ & $mAP$ \rule{0pt}{2.3ex}\rule[-0.9ex]{0pt}{0pt}\\
        \midrule
        $0$ & $80.9$ & $62.9$ \rule{0pt}{2.3ex}\\
        $1$ & $83.1$ & $66.0$ \rule{0pt}{2.3ex}\\
        $10$ & $85.3$ & $68.3$ \rule{0pt}{2.3ex}\\
        $20$ & $85.6$ & $70.2$ \rule{0pt}{2.3ex}\\
        $30$ & $\mathbf{86.8}$ & $\mathbf{71.7}$ \rule{0pt}{2.3ex}\\
        $40$ & $85.8$ & $70.7$ \rule{0pt}{2.1ex}\\
        \bottomrule
    \end{tabular*}
    \caption{Analysis of the queue size (in epochs) in the Data Sampling Module. The queue size of 0 indicates that the queue system is not used during sampling.  The model is trained on MSMT17-merged and tested on Market-1501.}
    \label{tab:queue-length-param}
\end{table}

\subsubsection{Parameters of the Relabeling Module}
 
In Tab.~\ref{tab:relabeling-params}, we analyze the parameters $\tau_{rel}$, $\tau_{remove}$, and $\tau_{merge}$, which control the refinement of pseudo-labels in the Relabeling Module. Optimal values $\tau_{rel}=0.6$ and $\tau_{remove}=0.5$ balance label refinement and noise filtering, enhancing pseudo-label quality and improving overall cross-dataset Re-ID performance. We do not set $\tau_{remove}$ to larger values since it should not exceed $\tau_{rel}$. The merging threshold $\tau_{merge}=0.8$ also ensures that similar PIDs are consolidated, further reducing noise and improving the consistency of pseudo-labels.

\subsubsection{Encoder Sensitivity Analysis}

In Tab.~\ref{tab:encoders-archs}, we analyze the impact of different encoder and momentum encoder architectures on the cross-dataset Re-ID performance. Although the backbone choice is not a hyperparameter of DynaMix, this experiment helps assess the robustness of our method to architectural changes. In this comparison, we keep all DynaMix components unchanged to ensure fair evaluation.

As shown in the results, increasing the model capacity from ViT-small to ViT-base indeed improves generalization performance. Moreover, when compared with the results in Tab.~\ref{tab:comparison-single}, even DynaMix with a ViT-small backbone surpasses other transformer-based methods such as TransMatcher and PAT, which employ larger ViT variants. This further confirms the effectiveness of the proposed training strategy and demonstrates that DynaMix is backbone-agnostic.

\begin{table}[t]
    \centering
    \footnotesize
    \begin{minipage}{0.23\textwidth}
        \centering
        \begin{tabular*}{\textwidth}{@{\hspace{8pt}\extracolsep{\fill}}c|*{2}{c}@{\hspace{4pt}}}
            \toprule
            $\tau_{rel}$ & $Rank_1$ & $mAP$ \rule{0pt}{2.3ex}\rule[-0.9ex]{0pt}{0pt}\\
            \midrule
            $0.4$ & $90.3$ & $76.6$ \rule{0pt}{2.3ex}\\
            $0.5$ & $90.5$ & $77.1$ \rule{0pt}{2.3ex}\\
            $0.6$ & $\mathbf{90.7}$ & $\mathbf{77.2}$ \rule{0pt}{2.3ex}\\
            $0.7$ & $89.0$ & $75.2$ \rule{0pt}{2.3ex}\\
            $0.8$ & $88.7$ & $74.3$ \rule{0pt}{2.1ex}\\
            \bottomrule
        \end{tabular*}
        \label{tab:rel-param}
    \end{minipage}%
    \hfill
    \begin{minipage}{0.23\textwidth}
        \centering
        \begin{minipage}{\textwidth}
            \centering
            \begin{tabular*}{\textwidth}{@{\hspace{5pt}\extracolsep{\fill}}c|*{2}{c}@{\hspace{5pt}}}
                \toprule
                $\tau_{remove}$ & $Rank_1$ & $mAP$ \rule{0pt}{2.3ex}\rule[-0.9ex]{0pt}{0pt}\\
                \midrule
                $0.3$ & $90.0$ & $76.2$ \rule{0pt}{2.3ex}\\
                $0.4$ & $90.2$ & $76.4$ \rule{0pt}{2.3ex}\\
                $0.5$ & $\mathbf{91.6}$ & $\mathbf{77.5}$ \rule{0pt}{2.1ex}\\
                \bottomrule
            \end{tabular*}
            \label{tab:remove-param}
        \end{minipage}\\[0.3cm] 
        \begin{minipage}{\textwidth}
            \centering
            \begin{tabular*}{\textwidth}{@{\hspace{5pt}\extracolsep{\fill}}c|*{2}{c}@{\hspace{3pt}}}
                \toprule
                $\tau_{merge}$ & $Rank_1$ & $mAP$ \rule{0pt}{2.3ex}\rule[-0.9ex]{0pt}{0pt}\\
                \midrule
                $0.7$ & $92.2$ & $80.6$ \rule{0pt}{2.3ex}\\
                $0.8$ & $\mathbf{92.6}$ & $\mathbf{80.8}$ \rule{0pt}{2.3ex}\\
                $0.9$ & $92.0$ & $79.3$ \rule{0pt}{2.1ex}\\
                \bottomrule
            \end{tabular*}
            \label{tab:merge-param}
        \end{minipage}
    \end{minipage}
    \caption{Analysis of $\tau_{rel}$, $\tau_{remove}$, and $\tau_{merge}$ in the Relabeling Module. The model is trained on MSMT17-merged and tested on Market-1501.}
    \label{tab:relabeling-params}
\end{table}

\begin{table}[t]
  \centering
  \footnotesize
  \begin{tabular*}{0.38\textwidth}{@{\hspace{1pt}}l|cc|cc}
    \toprule
    \multirow{2}{*}{Backbone} & \multicolumn{2}{c|}{CUHK03-NP} & \multicolumn{2}{@{\hspace{3pt}}c}{Market-1501}\\
    & $Rank_1$ & $mAP$ & $Rank_1$ & $mAP$ \rule{0pt}{2.3ex}\rule[-0.9ex]{0pt}{0pt}\\
    \midrule
    ViT-small & $52.0$ & $52.1$ & $89.9$ & $72.7$ \rule{0pt}{2.3ex}\\
    ViT-base & $\mathbf{60.9}$ & $\mathbf{61.0}$ & $\mathbf{92.6}$ & $\mathbf{80.8}$ \rule{0pt}{2.3ex}\\
    \bottomrule
  \end{tabular*}
  \caption{Comparison of different encoder architectures. The model is trained on MSMT17-merged and tested on Market-1501.}
  \label{tab:encoders-archs}
\end{table}

\subsubsection{Analysis of the Mini-Batch Size}

\begin{table}[t]
    \centering
    \footnotesize
    \begin{tabular*}{0.25\textwidth}{@{\hspace{6pt}\extracolsep{\fill}}c|*{2}{c}@{\hspace{3pt}}}
        \toprule
        $Size$ & $Rank_1$ & $mAP$ \rule{0pt}{2.3ex}\rule[-0.9ex]{0pt}{0pt}\\
        \midrule
        $32$ & $92.4$ & $80.5$ \rule{0pt}{2.3ex}\\
        $64$ & $\textbf{92.6}$ & $\textbf{80.8}$ \rule{0pt}{2.3ex}\\
        $128$ & $92.0$ & $79.4$ \rule{0pt}{2.3ex}\\
        $256$ & $91.9$ & $78.1$ \rule{0pt}{2.1ex}\\
        \bottomrule
    \end{tabular*}
    \caption{Analysis of the mini-batch size. The model is trained on MSMT17-merged and tested on Market-1501.}
    \label{tab:mini-batchsize-param}
\end{table}

We analyze the effect of the mini-batch size on cross-dataset Re-ID performance. In DynaMix, each mini-batch always contains an equal number of images from multi-camera and single-camera data, following the sampling strategy described in Sec.~\ref{sec:data-sampling-module}. By default, DynaMix uses a mini-batch size of $64$ (Sec.~\ref{sec:implementation-details}).

Tab.~\ref{tab:mini-batchsize-param} reports the impact of different mini-batch sizes on the generalization of our method. As the results show, the default choice provides the optimal balance. Increasing the mini-batch size beyond this point makes the optimization problem more difficult due to the heterogeneous nature of mixed data mini-batches, which leads to a degradation in performance. Conversely, reducing the mini-batch size also harms generalization, as it lowers the supervision signal. Overall, this analysis confirms that a moderate mini-batch size is most effective for maintaining both diversity and learning stability in DynaMix.

\section{Discussion}
\label{sec:discussion}

\subsection{How Does DynaMix Improve the Generalization?}

In real-world applications, the deployment environment is typically unknown: cameras, viewpoints, lighting, and background statistics vary drastically across locations. Therefore, strong generalization to unseen domains is a core requirement in person Re-ID. However, existing multi-camera datasets are relatively small (Tab.~\ref{tab:datasets}), which limits their stylistic diversity and prevents models from learning domain-invariant identity representations.

Large-scale single-camera datasets mitigate this limitation by providing images captured by thousands of cameras across diverse environments. Their stylistic and environmental variety can significantly improve generalization. However, using such data directly is challenging due to (1) noisy pseudo-labels, (2) the computational scale of millions of images, and (3) the strong distributional mismatch between multi-camera and single-camera data.

DynaMix addresses these challenges through its three modules: the Relabeling Module suppresses pseudo-label noise, the Efficient Centroids Module enables scalable and stable representation updates, and the Data Sampling Module constructs balanced mixed mini-batches that leverage the diversity of single-camera data effectively. As our experiments demonstrate, this combination substantially improves cross-dataset performance and, consequently, the generalization ability of Re-ID models by making better use of large-scale single-camera data.

\subsection{Synergy Between Modules is Everything}

The three modules in DynaMix are designed to complement one another, and their interaction is essential for achieving strong generalization. The Relabeling Module reduces pseudo-label noise, but its effect becomes significantly stronger when the Efficient Centroids Module provides stable identity representations across epochs. In turn, accurate centroid estimates allow the Data Sampling Module to construct informative mixed mini-batches, ensuring balanced supervision across heterogeneous data sources.

Moreover, by constructing balanced mixed data mini-batches, the Data Sampling Module enables the encoder to learn more robust features during each epoch. Consequently, the momentum encoder at the start of the next epoch produces more reliable embeddings, which directly improves the accuracy of pseudo-label refinement. This creates a natural iterative improvement cycle across epochs.

Thus, all components are mutually dependent and cannot function effectively in isolation. As demonstrated by our experiments (Sec.~\ref{sec:ablation}), their combination yields substantial improvements in cross-dataset performance, enabling DynaMix to fully exploit the diversity of large-scale single-camera data.

\section{Conclusion}
\label{sec:conclusion}

In this paper, we proposed DynaMix, a novel generalizable person Re-ID method that effectively combines manually labeled multi-camera and large-scale pseudo-labeled single-camera data. The strength of DynaMix lies in three core components: the Relabeling Module for dynamic refinement of noisy pseudo-labels, the Efficient Centroids Module for scalable identity representation updates, and the Data Sampling Module that carefully composes mixed data mini-batches. All components are specifically designed to operate efficiently at scale, enabling effective training on millions of images and hundreds of thousands of identities. Extensive experiments demonstrated that DynaMix achieves state-of-the-art performance across all cross-dataset and multi-source cross-dataset benchmarks. We believe that DynaMix sets a strong foundation for future research in generalizable person Re-ID and related visual recognition tasks.

\section{Acknowledgments}

This work was supported by the Ministry of Economic Development of the Russian Federation in accordance with the subsidy agreement (agreement identifier 000000C313925P4H0002; grant No 139-15-2025-012).


\begin{thebibliography}{00}


\bibitem{fu2021unsupervised}
  Dengpan Fu, Dongdong Chen, Jianmin Bao, Hao Yang, Lu Yuan, Lei Zhang, Houqiang Li, Dong Chen,  
  \textit{Unsupervised pre-training for person re-identification},  
  Proceedings of the IEEE/CVF Conference on Computer Vision and Pattern Recognition,  
  pp. 14750--14759, 2021.

\bibitem{fu2022large}
  Dengpan Fu, Dongdong Chen, Hao Yang, Jianmin Bao, Lu Yuan, Lei Zhang, Houqiang Li, Fang Wen, Dong Chen,  
  \textit{Large-Scale Pre-training for Person Re-identification with Noisy Labels},  
  arXiv preprint arXiv:2203.16533, 2022.

\bibitem{mamedov2023approaches}
  Timur Mamedov, Denis Kuplyakov, Anton Konushin,  
  \textit{Approaches to Improve the Quality of Person Re-Identification for Practical Use},  
  Sensors, vol. 23, no. 17, p. 7382, MDPI, 2023.

\bibitem{mamedov2025remix}
  Timur Mamedov, Anton Konushin, Vadim Konushin,  
  \textit{ReMix: Training Generalized Person Re-Identification on a Mixture of Data},  
  Winter Conference on Applications of Computer Vision (WACV),  
  pp. 8175--8185, February 2025.

\bibitem{prosser2010person}
  Bryan James Prosser, Wei-Shi Zheng, Shaogang Gong, Tao Xiang, Q. Mary et al.,  
  \textit{Person re-identification by support vector ranking},  
  BMVC, vol. 2, no. 5, p. 6, 2010.

\bibitem{zheng2012reidentification}
  Wei-Shi Zheng, Shaogang Gong, Tao Xiang,  
  \textit{Reidentification by relative distance comparison},  
  IEEE Transactions on Pattern Analysis and Machine Intelligence,  
  vol. 35, no. 3, pp. 653--668, 2012.

\bibitem{wang2018learning}
  Guanshuo Wang, Yufeng Yuan, Xiong Chen, Jiwei Li, Xi Zhou,  
  \textit{Learning discriminative features with multiple granularities for person re-identification},  
  Proceedings of the 26th ACM International Conference on Multimedia,  
  pp. 274--282, 2018.

\bibitem{luo2019bag}
  Hao Luo, Youzhi Gu, Xingyu Liao, Shenqi Lai, Wei Jiang,  
  \textit{Bag of tricks and a strong baseline for deep person re-identification},  
  Proceedings of the IEEE/CVF Conference on Computer Vision and Pattern Recognition Workshops,  
  2019.

\bibitem{ni2021flipreid}
  Xingyang Ni, Esa Rahtu,  
  \textit{FlipReID: Closing the Gap Between Training and Inference in Person Re-Identification},  
  9th European Workshop on Visual Information Processing (EUVIP),  
  pp. 1--6, IEEE, 2021.

\bibitem{he2021transreid}
  Shuting He, Hao Luo, Pichao Wang, Fan Wang, Hao Li, Wei Jiang,  
  \textit{TransReID: Transformer-based object re-identification},  
  Proceedings of the IEEE/CVF International Conference on Computer Vision,  
  pp. 15013--15022, 2021.

\bibitem{tan2022dynamic}
  Lei Tan, Pingyang Dai, Rongrong Ji, Yongjian Wu,  
  \textit{Dynamic prototype mask for occluded person re-identification},  
  Proceedings of the 30th ACM International Conference on Multimedia,  
  pp. 531--540, 2022.

\bibitem{zhu2022dual}
  Haowei Zhu, Wenjing Ke, Dong Li, Ji Liu, Lu Tian, Yi Shan,  
  \textit{Dual cross-attention learning for fine-grained visual categorization and object re-identification},  
  Proceedings of the IEEE/CVF Conference on Computer Vision and Pattern Recognition,  
  pp. 4692--4702, 2022.

\bibitem{zhu2022pass}
  Kuan Zhu, Haiyun Guo, Tianyi Yan, Yousong Zhu, Jinqiao Wang, Ming Tang,  
  \textit{PASS: Part-aware self-supervised pre-training for person re-identification},  
  European Conference on Computer Vision,  
  pp. 198--214, Springer, 2022.

\bibitem{zhang2023pha}
  Guiwei Zhang, Yongfei Zhang, Tianyu Zhang, Bo Li, Shiliang Pu,  
  \textit{PHA: Patch-wise high-frequency augmentation for transformer-based person re-identification},  
  Proceedings of the IEEE/CVF Conference on Computer Vision and Pattern Recognition,  
  pp. 14133--14142, 2023.

\bibitem{li2021diverse}
  Yulin Li, Jianfeng He, Tianzhu Zhang, Xiang Liu, Yongdong Zhang, Feng Wu,  
  \textit{Diverse part discovery: Occluded person re-identification with part-aware transformer},  
  Proceedings of the IEEE/CVF Conference on Computer Vision and Pattern Recognition,  
  pp. 2898--2907, 2021.

\bibitem{lai2021transformer}
  Shenqi Lai, Zhenhua Chai, Xiaolin Wei,  
  \textit{Transformer meets part model: Adaptive part division for person re-identification},  
  Proceedings of the IEEE/CVF International Conference on Computer Vision,  
  pp. 4150--4157, 2021.

\bibitem{jia2022learning}
  Mengxi Jia, Xinhua Cheng, Shijian Lu, Jian Zhang,  
  \textit{Learning disentangled representation implicitly via transformer for occluded person re-identification},  
  IEEE Transactions on Multimedia, vol. 25, pp. 1294--1305, 2022.

\bibitem{yan2022person}
  Pu Yan, Qingwei Tang, Jie Chen, Gang Wang, Yue Fang,  
  \textit{Person re-identification network based on weight-driven saliency hierarchical utilization},  
  Journal of Electronic Imaging,  
  vol. 31, no. 3, pp. 033014--033014, SPIE, 2022.

\bibitem{tang2022person}
  Qingwei Tang, Pu Yan, Jie Chen, Hui Shao, Fuyu Wang, Gang Wang,  
  \textit{Person re-identification based on multi-scale global feature and weight-driven part feature},  
  AI Communications,  
  vol. 35, no. 3, pp. 207--223, SAGE Publications, 2022.

\bibitem{li2023clip}
  Siyuan Li, Li Sun, Qingli Li,  
  \textit{CLIP-ReID: Exploiting vision-language model for image re-identification without concrete text labels},  
  Proceedings of the AAAI Conference on Artificial Intelligence,  
  vol. 37, no. 1, pp. 1405--1413, 2023.

\bibitem{jin2020style}
  Guiwei Zhang, Yongfei Zhang, Tianyu Zhang, Bo Li, Shiliang Pu,  
  \textit{PHA: Patch-wise high-frequency augmentation for transformer-based person re-identification},  
  Proceedings of the IEEE/CVF Conference on Computer Vision and Pattern Recognition,  
  pp. 14133--14142, 2023.

\bibitem{choi2021meta}
  Seokeon Choi, Taekyung Kim, Minki Jeong, Hyoungseob Park, Changick Kim,  
  \textit{Meta batch-instance normalization for generalizable person re-identification},  
  Proceedings of the IEEE/CVF Conference on Computer Vision and Pattern Recognition,  
  pp. 3425--3435, 2021.

\bibitem{jiao2022dynamically}
  Bingliang Jiao, Lingqiao Liu, Liying Gao, Guosheng Lin, Lu Yang, Shizhou Zhang, Peng Wang, Yanning Zhang,  
  \textit{Dynamically transformed instance normalization network for generalizable person re-identification},  
  European Conference on Computer Vision,  
  pp. 285--301, Springer, 2022.

\bibitem{ni2023part}
  Hao Ni, Yuke Li, Lianli Gao, Heng Tao Shen, Jingkuan Song,  
  \textit{Part-Aware Transformer for Generalizable Person Re-identification},  
  Proceedings of the IEEE/CVF International Conference on Computer Vision,  
  pp. 11280--11289, 2023.

\bibitem{zhou2019omni}
  Kaiyang Zhou, Yongxin Yang, Andrea Cavallaro, Tao Xiang,  
  \textit{Omni-scale feature learning for person re-identification},  
  Proceedings of the IEEE/CVF International Conference on Computer Vision,  
  pp. 3702--3712, 2019.

\bibitem{zhou2021learning}
  Kaiyang Zhou, Yongxin Yang, Andrea Cavallaro, Tao Xiang,  
  \textit{Learning generalisable omni-scale representations for person re-identification},  
  IEEE Transactions on Pattern Analysis and Machine Intelligence,  
  vol. 44, no. 9, pp. 5056--5069, 2021.

\bibitem{liao2021transmatcher}
  Shengcai Liao, Ling Shao,  
  \textit{Transmatcher: Deep image matching through transformers for generalizable person re-identification},  
  Advances in Neural Information Processing Systems,  
  vol. 34, pp. 1992--2003, 2021.

\bibitem{hu2024personvit}
  Bin Hu, Xinggang Wang, Wenyu Liu,  
  \textit{PersonViT: Large-scale Self-supervised Vision Transformer for Person Re-Identification},  
  arXiv preprint arXiv:2408.05398, 2024.

\bibitem{ding2024clothes}
  Yongkang Ding, Rui Mao, Guodong Du, Liyan Zhang,  
  \textit{Clothes-eraser: Clothing-aware controllable disentanglement for clothes-changing person re-identification},  
  Signal, Image and Video Processing,  
  vol. 18, no. 5, pp. 4337--4348, Springer, 2024.

\bibitem{ding2025decoupling}
  Yongkang Ding, Xiaoyin Wang, Hao Yuan, Meina Qu, Xiangzhou Jian,  
  \textit{Decoupling feature-driven and multimodal fusion attention for clothing-changing person re-identification},  
  Artificial Intelligence Review,  
  vol. 58, no. 8, p. 241, Springer, 2025.

\bibitem{ding2025person}
  Yongkang Ding, Yuqing Wu, Chenwei Wu, Meina Qu, Liyan Zhang,  
  \textit{Person Parsing-Driven and Text-Guided for Cloth-Changing Person Re-Identification},  
  IEEE Internet of Things Journal,  
  IEEE, 2025.

\bibitem{pang2023cross}
  Zhiqi Pang, Chunyu Wang, Lingling Zhao, Yang Liu, Gaurav Sharma,  
  \textit{Cross-modality hierarchical clustering and refinement for unsupervised visible-infrared person re-identification},  
  IEEE Transactions on Circuits and Systems for Video Technology,  
  vol. 34, no. 4, pp. 2706--2718, IEEE, 2023.

\bibitem{ren2024implicit}
  Kaijie Ren, Lei Zhang,  
  \textit{Implicit discriminative knowledge learning for visible-infrared person re-identification},  
  Proceedings of the IEEE/CVF Conference on Computer Vision and Pattern Recognition,  
  pp. 393--402, 2024.

\bibitem{tang2024visible}
  Qingwei Tang, Pu Yan, Wei Sun,  
  \textit{Visible-infrared person re-identification employing style-supervision and content-supervision},  
  The Visual Computer,  
  vol. 40, no. 4, pp. 2443--2456, Springer, 2024.

\bibitem{zhang2021deep}
  Guoqing Zhang, Yu Ge, Zhicheng Dong, Hao Wang, Yuhui Zheng, Shengyong Chen,  
  \textit{Deep high-resolution representation learning for cross-resolution person re-identification},  
  IEEE Transactions on Image Processing,  
  vol. 30, pp. 8913--8925, IEEE, 2021.

\bibitem{pang2024dual}
  Zhiqi Pang, Lingling Zhao, Chunyu Wang,  
  \textit{Dual-resolution fusion modeling for unsupervised cross-resolution person re-identification},  
  Proceedings of the 32nd ACM International Conference on Multimedia,  
  pp. 4063--4072, 2024.

\bibitem{wojke2017simple}
  Nicolai Wojke, Alex Bewley, Dietrich Paulus,  
  \textit{Simple online and realtime tracking with a deep association metric},  
  Proceedings of the IEEE International Conference on Image Processing (ICIP),  
  pp. 3645--3649, IEEE, 2017.

\bibitem{mamedov2021queue}
  Timur Mamedov, Denis Kuplyakov, Anton Konushin,  
  \textit{Queue Waiting Time Estimation Using Person Re-identification by Upper Body},  
  Proceedings of the 31st International Conference on Computer Graphics and Machine Vision,  
  pp. 27--30, 2021.

\bibitem{zhang2022bytetrack}
  Yifu Zhang, Peize Sun, Yi Jiang, Dongdong Yu, Fucheng Weng, Zehuan Yuan, Ping Luo, Wenyu Liu, Xinggang Wang,  
  \textit{ByteTrack: Multi-object tracking by associating every detection box},  
  European Conference on Computer Vision,  
  pp. 1--21, Springer, 2022.

\bibitem{mamedov2022video}
  T.Z. Mamedov, D. A. Kuplyakov, A. S. Konushin,  
  \textit{Video Analytics Using Detection on Sparse Frames},  
  Programming and Computer Software,  
  vol. 48, no. 3, pp. 155--163, Springer, 2022.

\bibitem{kuhn1955hungarian}
  Harold W. Kuhn,  
  \textit{The Hungarian method for the assignment problem},  
  Naval Research Logistics Quarterly,  
  vol. 2, no. 1--2, pp. 83--97, Wiley, 1955.

\bibitem{li2014deepreid}
  Wei Li, Rui Zhao, Tong Xiao, Xiaogang Wang,  
  \textit{DeepReID: Deep filter pairing neural network for person re-identification},  
  Proceedings of the IEEE Conference on Computer Vision and Pattern Recognition,  
  pp. 152--159, 2014.

\bibitem{zheng2015scalable}
  Liang Zheng, Liyue Shen, Lu Tian, Shengjin Wang, Jingdong Wang, Qi Tian,  
  \textit{Scalable person re-identification: A benchmark},  
  Proceedings of the IEEE International Conference on Computer Vision,  
  pp. 1116--1124, 2015.

\bibitem{wei2018person}
  Longhui Wei, Shiliang Zhang, Wen Gao, Qi Tian,  
  \textit{Person transfer GAN to bridge domain gap for person re-identification},  
  Proceedings of the IEEE Conference on Computer Vision and Pattern Recognition,  
  pp. 79--88, 2018.

\bibitem{ristani2016performance}
  Ergys Ristani, Francesco Solera, Roger Zou, Rita Cucchiara, Carlo Tomasi,  
  \textit{Performance measures and a data set for multi-target, multi-camera tracking},  
  European Conference on Computer Vision,  
  pp. 17--35, Springer, 2016.

\bibitem{wang2020surpassing}
  Yanan Wang, Shengcai Liao, Ling Shao,  
  \textit{Surpassing real-world source training data: Random 3D characters for generalizable person re-identification},  
  Proceedings of the 28th ACM International Conference on Multimedia,  
  pp. 3422--3430, 2020.

\bibitem{dosovitskiy2020image}
  Alexey Dosovitskiy et al.,  
  \textit{An image is worth 16x16 words: Transformers for image recognition at scale},  
  arXiv preprint arXiv:2010.11929, 2020.

\bibitem{paszke2019pytorch}
  Adam Paszke, Sam Gross, Francisco Massa, Adam Lerer, James Bradbury, Gregory Chanan, Trevor Killeen, Zeming Lin, Natalia Gimelshein, Luca Antiga, Alban Desmaison, Andreas Köpf, Edward Yang, Zach DeVito, Martin Raison, Alykhan Tejani, Sasank Chilamkurthy, Benoit Steiner, Lu Fang, Junjie Bai, Soumith Chintala,  
  \textit{PyTorch: An imperative style, high-performance deep learning library},  
  Proceedings of the 33rd International Conference on Neural Information Processing Systems (NeurIPS),  
  Curran Associates Inc., Red Hook, NY, USA,  
  Article no. 721, pp. 12, 2019.

\bibitem{liao2020interpretable}
  Shengcai Liao, Ling Shao,  
  \textit{Interpretable and generalizable person re-identification with query-adaptive convolution and temporal lifting},  
  European Conference on Computer Vision (ECCV),  
  pp. 456--474, Springer, 2020.

\bibitem{liao2022graph}
  Shengcai Liao, Ling Shao,  
  \textit{Graph sampling based deep metric learning for generalizable person re-identification},  
  Proceedings of the IEEE/CVF Conference on Computer Vision and Pattern Recognition,  
  pp. 7359--7368, 2022.

\bibitem{peng2024invariance}
  Wanru Peng, Houjin Chen, Yanfeng Li, Jia Sun,  
  \textit{Invariance Learning under Uncertainty for Single Domain Generalization Person Re-Identification},  
  IEEE Transactions on Instrumentation and Measurement, 2024.

\bibitem{li2025unleashing}
  Jiachen Li, Xiaojin Gong,  
  \textit{Unleashing the Potential of Pre-Trained Diffusion Models for Generalizable Person Re-Identification},  
  Sensors, vol. 25, no. 2, p. 552, 2025.

\bibitem{zhao2024clip}
  Huazhong Zhao, Lei Qi, Xin Geng,  
  \textit{CLIP-DFGS: A Hard Sample Mining Method for CLIP in Generalizable Person Re-Identification},  
  ACM Transactions on Multimedia Computing, Communications and Applications,  
  vol. 21, no. 1, pp. 1--20, 2024.

\bibitem{qi2022novel}
  Lei Qi, Lei Wang, Yinghuan Shi, Xin Geng,  
  \textit{A novel mix-normalization method for generalizable multi-source person re-identification},  
  IEEE Transactions on Multimedia, 2022.

\bibitem{xu2021meta}
  Boqiang Xu, Jian Liang, Lingxiao He, Zhenan Sun,  
  \textit{Meta: Mimicking embedding via others’ aggregation for generalizable person re-identification},  
  Proceedings of the European Conference on Computer Vision (ECCV), 2022.

\bibitem{tan2023style}
  Wentao Tan, Changxing Ding, Pengfei Wang, Mingming Gong, Kui Jia,  
  \textit{Style interleaved learning for generalizable person re-identification},  
  IEEE Transactions on Multimedia, 2023.

\bibitem{dou2023identity}
  Zhaopeng Dou, Zhongdao Wang, Yali Li, Shengjin Wang,  
  \textit{Identity-seeking self-supervised representation learning for generalizable person re-identification},  
  Proceedings of the IEEE/CVF International Conference on Computer Vision,  
  pp. 15847--15858, 2023.

\bibitem{ang2024unified}
  Eugene P. W. Ang, Shan Lin, Alex C. Kot,  
  \textit{A unified deep semantic expansion framework for domain-generalized person re-identification},  
  Neurocomputing, vol. 600, p. 128120, Elsevier, 2024.

\bibitem{cho2024generalizable}
  Yoonki Cho, Jaeyoon Kim, Woo Jae Kim, Junsik Jung, Sung-eui Yoon,  
  \textit{Generalizable Person Re-identification via Balancing Alignment and Uniformity},  
  arXiv preprint arXiv:2411.11471, 2024.

\bibitem{he2016deep}
  Kaiming He, Xiangyu Zhang, Shaoqing Ren, Jian Sun,  
  \textit{Deep residual learning for image recognition},  
  Proceedings of the IEEE Conference on Computer Vision and Pattern Recognition,  
  pp. 770--778, 2016.

\bibitem{pan2018two}
  Xingang Pan, Ping Luo, Jianping Shi, Xiaoou Tang,  
  \textit{Two at once: Enhancing learning and generalization capacities via IBN-Net},  
  Proceedings of the European Conference on Computer Vision (ECCV),  
  pp. 464--479, 2018.

\end{thebibliography}
\end{document}